\documentclass[letterpaper, 10 pt, conference]{ieeeconf}

\IEEEoverridecommandlockouts
\usepackage{amsmath}
\usepackage{amssymb}
\usepackage{booktabs}
\usepackage{cite}
\usepackage{wrapfig}
\usepackage{graphicx}
\usepackage{multirow}
\usepackage{url}
\usepackage{xcolor}

\newcommand{\TTD}{\mathrm{TTD}}
\newcommand{\Em}{E_{\mathrm{m}}}
\newcommand{\cmark}{\textcolor{green!50!black}{\checkmark}}
\newcommand{\xmark}{\textcolor{red!75!black}{$\times$}}
\newcommand{\pmark}{\textcolor{orange!90!black}{$\blacktriangle$}}

\makeatletter
\def\thebibliography#1{\section*{References}%
    \addcontentsline{toc}{section}{References}%
    \fontsize{8.0pt}{8.0pt}\selectfont \vskip 0.1\baselineskip%
    \list{\@biblabel{\@arabic\c@enumiv}}%
    {\settowidth\labelwidth{\@biblabel{#1}}%
    \leftmargin\labelwidth
    \advance\leftmargin\labelsep\relax
    \itemsep 0pt\relax%
    \usecounter{enumiv}%
    \let\p@enumiv\@empty
    \renewcommand\theenumiv{\@arabic\c@enumiv}}%
    \let\@IEEElatexbibitem\bibitem%
    \def\bibitem{\@IEEEbibitemprefix\@IEEElatexbibitem}%
    \def\newblock{\hskip .11em plus .33em minus .07em}%
    \sloppy\clubpenalty4000\widowpenalty4000\interlinepenalty500%
    \sfcode`\.=1000\relax}
\makeatother

\title{\LARGE \bf
Energy-Aware Open-Vocabulary Detection Scheduling for Low-Power Robotic Autonomy
}

\author{Abdul Basit and Muhammad Shafique%
\thanks{Abdul Basit and Muhammad Shafique are with eBRAIN Lab, New York University Abu Dhabi, Abu Dhabi, UAE.
{\tt\footnotesize abdul.basit@nyu.edu, muhammad.shafique@nyu.edu}}%
\thanks{This work was supported in part by the NYUAD Center for Cyber Security (CCS), funded by Tamkeen under the NYUAD Research Institute Award G1104.}%
}

\begin{document}

\maketitle
\thispagestyle{empty}
\pagestyle{empty}
\bstctlcite{IEEEexample:BSTcontrol}

\begin{abstract}
Open-vocabulary object detection (OVD) lets robots search for run-time object queries, but fixed-rate inference spends embedded compute on low-value frames during idle motion, map revisits, or stable viewpoints. This paper presents an energy-aware OVD scheduler for low-power mobile robots. The scheduler uses odometry, map change, and detector confidence to switch among low-, mid-, and high-power YOLO-World-S modes on a Jetson Orin Nano. We evaluate the energy-quality trade-off on an AgileX LIMO using 40 real navigation trials: two target objects, five placements per target-policy pair, and four perception policies. Across bottle and cup trials, the scheduler records 8/10 mission success compared with 7/10 for each fixed-rate baseline, while reducing mean Jetson input power (a Jetson-side power-draw metric) by 5.2\% relative to an 8 Hz high-rate baseline and lowering achieved OVD call rate by 34.0\%. It does not minimize mission-window Jetson input energy because longer missions can erase instantaneous power savings. A separate 60 s stationary microbenchmark isolates the perception workload: the scheduler lowers mean power from 8.045 W to 7.301 W and energy from 484.7 J to 438.1 J relative to the 8 Hz baseline. These results support robot-state-aware OVD scheduling as a promising power-management mechanism with competitive task reliability, while showing that mission-window Jetson input energy must be read together with task progress and runtime.
\end{abstract}

\section{Introduction}

Open-vocabulary object detection (OVD) lets mobile robots ground run-time object queries, such as ``find the bottle near the doorway,'' in their camera stream instead of searching only fixed trained labels. This capability is enabled by vision-language models and open-vocabulary detection/grounding systems, including CLIP~\cite{radford2021learningtransferablevisualmodels}, OWL-ViT~\cite{minderer2022simpleopenvocabularyobjectdetection}, GLIP~\cite{li2022glip}, Grounding DINO~\cite{liu2024groundingdinomarryingdino}, Detic~\cite{zhou2022detic}, and YOLO-World~\cite{cheng2024yoloworldrealtimeopenvocabularyobject}. In robotics, similar representations support language-grounded maps, open-vocabulary scene graphs, and object-centric navigation~\cite{huang2023vlmaps,werby2024scenegraphs,gu2024conceptgraphs,shah2023lmnav}.

The deployment problem is that OVD is expensive on embedded robots. A low-power robot must share a small battery across the embedded GPU, RGB-D camera, LiDAR, wireless stack, controller, and drive base. Mean Jetson input power is therefore a practical Jetson-side power-draw metric, especially for robots that keep monitoring or waiting after a target event. A fixed high detector rate gives fast response but wastes power during idle motion, map revisits, or low-information views. A fixed low-rate or quantized detector saves power but can delay target acceptance or miss short visual opportunities. The right operating point changes during the mission.

This motivates mission-level perception scheduling: each inference should be worth its energy at the current robot state. We ask whether an embedded robot can reduce Jetson input power for OVD while maintaining mission completion by scheduling detector rate from robot state. Our hypothesis is that low-value frames can be sampled sparsely, while candidate-target phases need high-rate response. A stationary frame in a stable map has lower marginal value than a frame near a plausible target; once a target appears, aggressive throttling is risky.

Existing work covers pieces of the problem but not the full intersection. OVD papers emphasize zero-shot recognition, language-grounded navigation emphasizes semantic goal reaching, and adaptive video analytics emphasizes throughput or resource budgets for passive streams. Our focus is more targeted: deciding when an embedded robot should spend OVD compute during search. As summarized in Table~\ref{tab:related_coverage}, prior works generally do not combine OVD, robot-state-aware adaptation, Jetson input-power/energy accounting, and task-level metrics such as success rate and time-to-detect (TTD).

We present an energy-aware OVD scheduler that treats detector invocations as mission-level computational resources. The scheduler uses odometry, map changes, and detector confidence to adaptively switch among low-, mid-, and high-power operating modes. We evaluate YOLO-World-S on a Jetson Orin Nano mobile robot against fixed-rate 8 Hz, 5 Hz, and 2 Hz baselines across bottle and cup search tasks. Fig.~\ref{fig:method} illustrates the overall runtime flow. 

Our main contributions are:
\begin{itemize}
    \item We formulate energy-aware OVD as a mission-level scheduling problem that trades Jetson input power and detector invocation rate against search success and TTD, rather than only reducing per-frame inference cost.
    \item We design a finite-state runtime that switches OVD rate from robot-native signals with explicit thresholds and hysteresis, including a low-power configuration for low-value frames.
    \item We implement the scheduler with YOLO-World-S on an AgileX LIMO equipped with a Jetson Orin Nano and compare it against fixed-rate high-, medium-, and low-rate baselines.
    \item We report a real-robot evaluation with 40 navigation trials across two queried objects and a separate stationary microbenchmark that isolates perception power.
    \item We show empirically that lower mean power does not necessarily imply lower mission-window Jetson input energy, because slower or less reliable perception can increase navigation time.
\end{itemize}

The main finding is that lower instantaneous detector power is only one part of the robot objective. The scheduler reduces mean Jetson input power relative to the high-rate baseline and shows slightly higher aggregate success in the evaluation logs, but mission-window Jetson input energy still depends strongly on runtime. We therefore report power, energy, success, TTD, and time-domain traces rather than optimizing energy in isolation.

\section{Related Work}

Table~\ref{tab:related_coverage} positions representative prior work by this paper's evaluation axes. The comparison is qualitative: the proposed scheduler builds on OVD models but studies their coupling with robot-state-aware scheduling, Jetson input-power/energy accounting, and task-quality metrics.

\begin{table}[ht]
\centering
\scriptsize
\setlength{\tabcolsep}{1.8pt}
\caption{Qualitative coverage of representative related work.}
\label{tab:related_coverage}
\resizebox{\columnwidth}{!}{%
\begin{tabular}{@{}lccccc@{}}
\toprule
\textbf{Work} & \textbf{OVD} & \textbf{Robot} & \textbf{Adapt.} & \textbf{$R/E$} & \textbf{Task}\\
\midrule
OWL-ViT/GLIP/GDINO~\cite{minderer2022simpleopenvocabularyobjectdetection,li2022glip,liu2024groundingdinomarryingdino} & \cmark & \xmark & \xmark & \xmark & \xmark\\
Detic/YOLO-World~\cite{zhou2022detic,cheng2024yoloworldrealtimeopenvocabularyobject} & \cmark & \xmark & \xmark & \pmark & \xmark\\
VLMaps/LM-Nav/ROVER~\cite{huang2023vlmaps,shah2023lmnav,basit2025rover} & \pmark & \cmark & \xmark & \xmark & \pmark\\
OV scene graphs~\cite{werby2024scenegraphs,gu2024conceptgraphs} & \cmark & \pmark & \xmark & \xmark & \pmark\\
Active/NBV perception~\cite{bajcsy1988active,korbach2021nextbestview} & \xmark & \cmark & \cmark & \pmark & \cmark\\
NoScope/Chameleon~\cite{kang2017noscope,jiang2018chameleon} & \xmark & \xmark & \cmark & \pmark & \xmark\\
ApproxDet~\cite{xu2020approxdet} & \xmark & \xmark & \cmark & \pmark & \xmark\\
\textbf{Proposed scheduler} & \cmark & \cmark & \cmark & \cmark & \cmark\\
\bottomrule
\end{tabular}
}

\begin{minipage}{0.98\columnwidth}
\vspace{0.5mm}
\scriptsize\emph{Note:} Robot = robot task evaluation; Adapt. = runtime adaptation; $R/E$ = resource/energy metric; Task = success/TTD. \cmark = covered, \pmark = partial/indirect, \xmark = not a primary focus.
\end{minipage}
\end{table}

Modern OVD systems inherit generalization from image-text pretraining. CLIP~\cite{radford2021learningtransferablevisualmodels} introduced transferable vision-language representations, and OWL-ViT~\cite{minderer2022simpleopenvocabularyobjectdetection}, GLIP~\cite{li2022glip}, RegionCLIP~\cite{zhong2022regionclip}, Detic~\cite{zhou2022detic}, Grounding DINO~\cite{liu2024groundingdinomarryingdino}, and YOLO-World~\cite{cheng2024yoloworldrealtimeopenvocabularyobject} adapt them to region-level localization. These works emphasize recognition accuracy, vocabulary generalization, and frame latency, but rarely ask how often an embedded robot should invoke OVD during a mission.

Open-vocabulary perception also supports semantic robot behavior. VLMaps~\cite{huang2023vlmaps}, LM-Nav~\cite{shah2023lmnav}, open-vocabulary scene graphs~\cite{werby2024scenegraphs}, ConceptGraphs~\cite{gu2024conceptgraphs}, ROVER~\cite{basit2025rover}, and RoboAuditor~\cite{cai2023roboauditor} use language-grounded perception for navigation, mapping, planning, object search, or inspection. These systems show the utility of open vocabularies for robots, but not the scheduling and energy cost of the detector.

Efficient inference and adaptive perception provide the closest systems context. Approximate video analytics adapts resolution, model size, or sampling rate for passive camera workloads~\cite{kang2017noscope,jiang2018chameleon,xu2020approxdet}; edge systems partition or accelerate neural computation~\cite{kang2017neurosurgeon,han2016eie,chen2016eyeriss}; continual-learning surveys highlight deployment adaptation challenges in autonomous systems~\cite{shaheen2022continual}; and active/NBV perception decides where or when to sense using task state or information gain~\cite{bajcsy1988active,korbach2021nextbestview}. In contrast, a mobile OVD robot has mission-native signals such as odometry, map change, and target confidence, and a missed target can change task outcome. These systems are complementary but optimize frame-level or edge-cloud decisions, whereas our evaluation couples OVD scheduling to closed-loop success, TTD, and mission-window power. This paper targets that intersection by treating OVD calls as mission resources and evaluating robot-state-aware scheduling with Jetson input energy, success, and TTD.

\section{Methodology}

\subsection{Problem Formulation}


We consider a differential-drive mobile robot performing language-guided object search using an embedded open-vocabulary detector (OVD). For each mission, the robot is provided with a target query \(q\) and autonomously searches the environment until either the task is completed or a mission duration \(\mathcal{T}_{\max}\) is reached. Perception and navigation operate continuously throughout the mission, while detector execution is governed by the proposed energy-aware scheduling policy. We define TTD as the elapsed time from mission start to the first detection of the queried object; this event is used as an evaluation metric and does not terminate the mission. At time \(t\), the perception runtime selects \(a(t) = \left(f(t), c_{\mathrm{req}}(t)\right)\), where \(f(t)\) is the requested OVD rate and \(c_{\mathrm{req}}(t)\in\{\mathrm{FP16},\mathrm{INT8}\}\) is the requested inference mode. The mission-window Jetson input energy is

\begin{equation}
    \Em = \int_{0}^{\mathcal{T}} P_{\mathrm{jetson}}(t)\,dt,
\end{equation}
where $\mathcal{T}\leq\mathcal{T}_{\max}$ is the realized duration. Because energy alone can be minimized by never running the detector, we measure search quality using success rate and time-to-detect,
\begin{equation}
    \TTD = t_{\mathrm{first\;conf}} - t_{\mathrm{start}} .
\end{equation}
where $t_{\mathrm{first\;conf}}$ is the first target detection above $\tau_{\mathrm{conf}}$. TTD is a perception-timing metric, not the success label. Aggregate TTD is reported over successful runs only; failures remain in success counts, runtime, and mission-window Jetson input energy.

The design objective is to reduce Jetson input power and detector workload while maintaining success and acceptable TTD relative to a high-rate detector baseline. We also report $\Em$ to capture the effect of mission duration. Unlike offline detector compression, the scheduler is conditioned on robot state and spends computation unevenly across the mission.

\begin{figure*}[ht]
    \centering
    \includegraphics[width=0.94\textwidth]{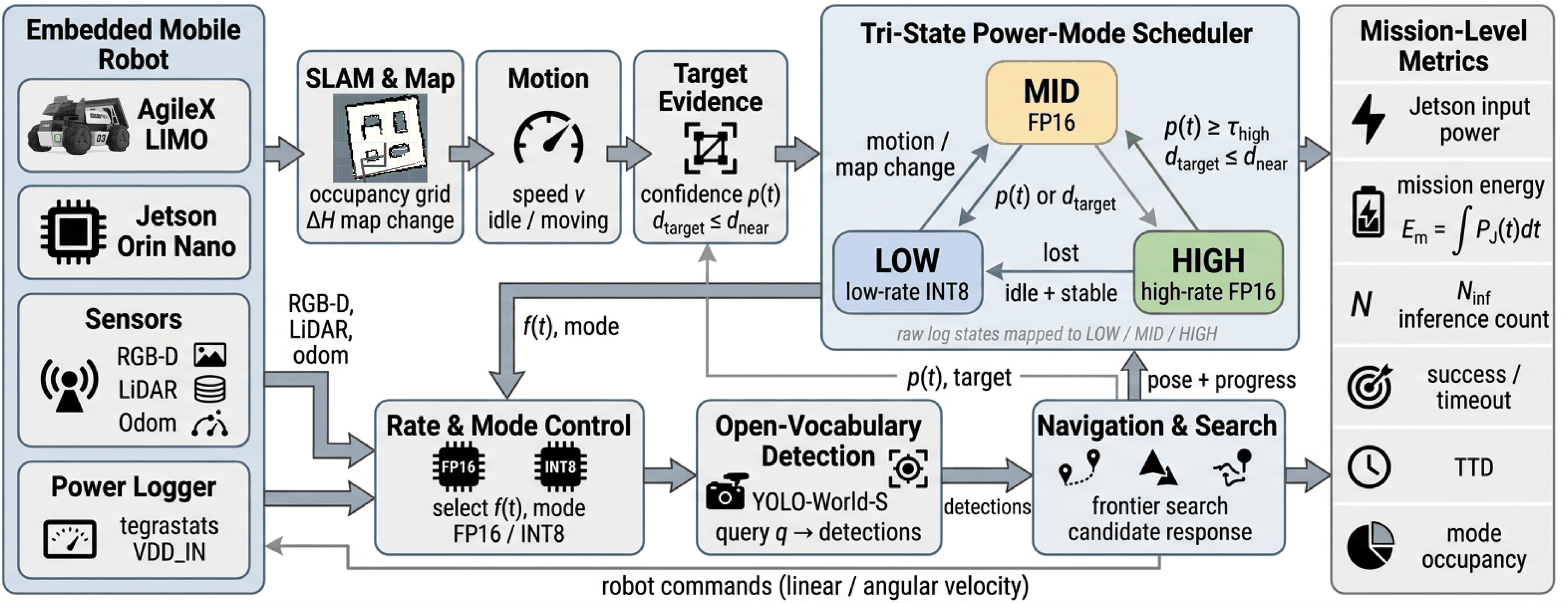}
    \caption{Proposed energy-aware OVD runtime. Robot sensing and mapping produce motion, map-change, confidence, and target-distance signals. The tri-state scheduler maps these signals to LOW/MID/HIGH detector modes, while the logging pipeline records Jetson input power, mission-window energy, inference count, success, TTD, and mode occupancy.}
    \label{fig:method}
\end{figure*}

\subsection{System Overview}

Fig.~\ref{fig:method} shows the runtime. The robot runs ROS Noetic with GMapping and \texttt{move\_base}. The OVD node receives RGB frames from an Astra/Dabai U3 RGB-D camera and evaluates YOLO-World-S for the queried class. Requested rates are upper bounds: the camera continues streaming, but frames reach the detector only when the scheduler permits a new OVD call. A state monitor subscribes to odometry, the occupancy grid, and detector outputs. A logger records scheduler events, poses, detector calls, mission outcomes, and Jetson power from \texttt{tegrastats}.

The scheduler uses three robot-state signals:
\begin{itemize}
    \item \textbf{Motion}: linear speed $v(t)$ from odometry, used to identify idle or nearly idle intervals.
    \item \textbf{Map change}: normalized occupancy-grid entropy change $\Delta H(t)$, used as a proxy for whether new space is being observed.
    \item \textbf{Detection confidence and target proximity}: best target confidence $p(t)$ from the OVD node and proximity to a recent target estimate, used to enter a high-rate candidate-response mode.
\end{itemize}

For an occupancy grid window $\Omega_t$, normalized map entropy is computed as
\begin{equation}
    \begin{aligned}
    H(t) &= \frac{1}{|\Omega_t|\log 2}\sum_{i\in\Omega_t} h_i(t),\\
    h_i(t) &= -p_i(t)\log p_i(t)
    -(1-p_i(t))\log(1-p_i(t)).
    \end{aligned}
\end{equation}
Unknown cells use $p_i=0.5$, and known cells use normalized occupancy probability clipped to $[\epsilon,1-\epsilon]$. We use a $4.0\,\mathrm{m}\times4.0\,\mathrm{m}$ robot-centered window; if successive GMapping maps differ in extent, $\Delta H$ uses overlapping indexed cells. The scheduler updates at $r_s=20$ Hz, so $\Delta H(t)=|H(t)-H(t-\Delta t_s)|$ with $\Delta t_s=1/r_s$. Detector confidence $p(t)$ updates after each OVD call. Only fresh detector output can satisfy the confidence branch of $M\rightarrow H$; the distance branch uses an unexpired target estimate. The estimate comes from RGB-D box-center back-projection to the map frame, LiDAR snapping within $0.3$ m, and exponential filtering with $\alpha=0.3$; after 30 s without update, $d_{\mathrm{target}}=\infty$.

\subsection{Tri-State Scheduling Policy}

The proposed scheduler uses a finite-state policy with hysteresis. Each state has a fixed target rate and preferred inference mode:
\begin{itemize}
    \item \textbf{LOW}: requested INT8. Used after sustained evidence that the robot is nearly idle and the local map is stable. This state samples frames sparsely to retain target awareness without continuous high-rate detection.
    \item \textbf{MID}: FP16. Used during normal navigation and mapping, where new viewpoints arrive but there is no confident target.
    \item \textbf{HIGH}: FP16. Used when a target candidate is detected or the robot is near a recent target estimate, prioritizing high-rate response over energy.
\end{itemize}
HIGH denotes a high-power candidate-response state rather than a distinct long-term tracking mode. The scheduler operates deterministically: transitions to HIGH are triggered by moderate evidence of the target, whereas entry into LOW is permitted only after sustained inactivity and persistent evidence of map stability.

For compactness, let $L$, $M$, and $H$ denote LOW, MID, and HIGH. The transition rules are:
\begin{align}
L\rightarrow M &: v>v_{\mathrm{idle}} \;\vee\; \Delta H>\tau_H,\\
M\rightarrow H &: p\geq\tau_{\mathrm{high}}\;\vee\;d_{\mathrm{target}}\leq d_{\mathrm{near}},\\
H\rightarrow M &: p<\tau_{\mathrm{hyst}}\;\text{for}\;K_{\mathrm{lost}}\;\text{evals.},\\
M\rightarrow L &: v<v_{\mathrm{idle}}\;\wedge\;\Delta H<\tau_H\;\text{for}\;T_{\mathrm{low}}.
\end{align}

Table~\ref{tab:thresholds} lists parameters used for all reported runs. They are fixed rather than tuned per trajectory: the system enters HIGH at moderate confidence but requires persistent low confidence before leaving.

\begin{table}[ht]
\centering
\footnotesize
\caption{Scheduler parameters used in the reported experiments.}
\label{tab:thresholds}
\begin{tabular}{@{}lll@{}}
\toprule
\textbf{Parameter} & \textbf{Value} & \textbf{Purpose}\\
\midrule
$r_s$ & 20 Hz & Scheduler update rate\\
$f_L/f_M/f_H$ & 1.5/4/8 Hz & Requested OVD rates\\
$\ell_{\Omega}$ & 4.0 m & Local grid-window side\\
$v_{\mathrm{idle}}$ & 0.03 m/s & Idle-motion threshold\\
$\tau_H$ & 0.02 & Map-change threshold\\
$T_{\mathrm{low}}$ & 2.0 s & Stable interval before LOW\\
$\tau_{\mathrm{high}}$ & 0.25 & Confidence to enter HIGH\\
$\tau_{\mathrm{conf}}$ & 0.25 & Detection acceptance threshold\\
$\tau_{\mathrm{int8}}$ & 0.20 & INT8 detector threshold\\
$\tau_{\mathrm{hyst}}$ & 0.20 & Confidence to leave HIGH\\
$d_{\mathrm{near}}$ & 2.0 m & Near-target HIGH trigger\\
$K_{\mathrm{lost}}$ & 5 evals. & HIGH hysteresis window\\
\bottomrule
\end{tabular}
\end{table}

\subsection{Low-Cost Inference Mode}


The perception runtime maintains both the high-rate YOLO-World-S inference path and the INT8 TensorRT engine in memory, selecting the appropriate execution mode according to the scheduler state. INT8 inference is used exclusively in LOW mode, where frames are processed at reduced precision to support coarse target awareness with lower computational cost. MID and HIGH modes use the FP16 inference path to provide higher-fidelity localization and response. The INT8 decision threshold, \(\tau_{\mathrm{int8}}\), is applied only within the LOW-mode detector, whereas mission-level detection decisions and time-to-detection (TTD) are evaluated using the common confidence threshold \(\tau_{\mathrm{conf}}\). All reported experiments were conducted with the INT8 engine available and without precision fallback.

\subsection{Metrics}
\label{sec:methodology_metrics}

For mission runs we log success, TTD, runtime, path length, OVD invocations $N_{\mathrm{inf}}$, average OVD rate $N_{\mathrm{inf}}/\mathcal{T}$, state occupancy, mean Jetson input power, and mission-window energy. State occupancy is the realized time fraction in LOW/MID/HIGH using the corresponding raw labels. Power is scoped to the Jetson module, not the full robot battery: the analysis uses \texttt{tegrastats} VDD\_IN \texttt{total\_W} at a 50 ms requested sampling interval, with GPU and SoC rails used only as diagnostics. For every reported trial, the power trace is aligned to mission start/end wall time and energy is integrated by a trapezoidal sum over valid in-window samples. Failed missions remain in all-run averages because they consume real time and power.

\section{Experimental Setup}

\subsection{Robot Platform}

\begin{figure}[ht]
  \centering
  \includegraphics[width=0.88\columnwidth]{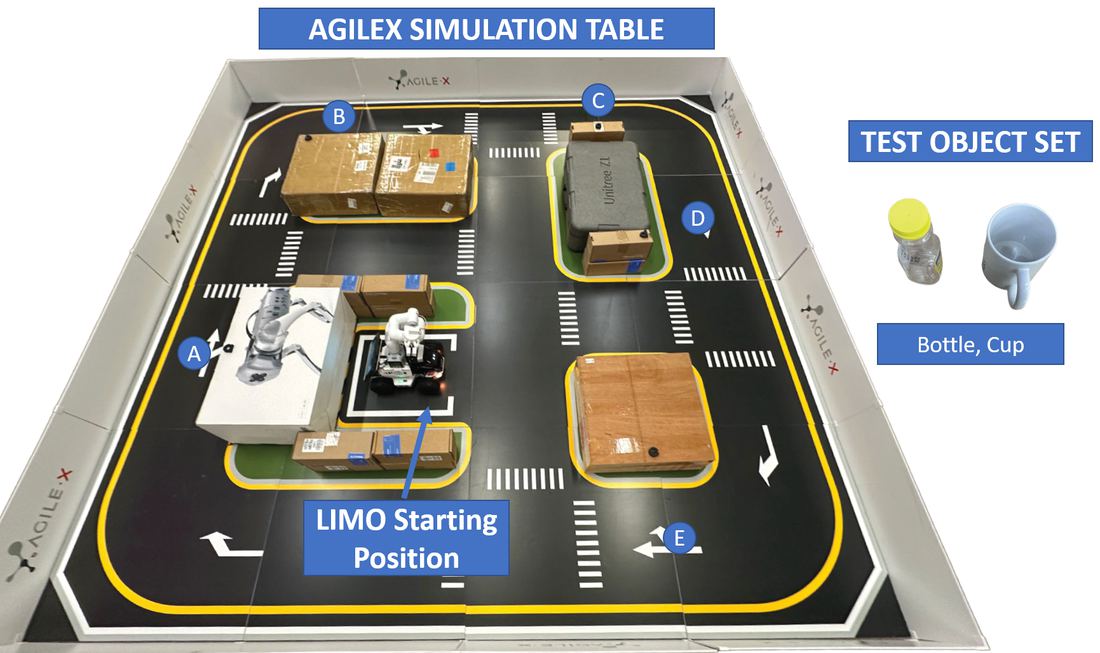}
  \caption{Real-robot arena with fixed LIMO start pose and bottle/cup target placements reused across policies.}
  \label{fig:exp_setup}
\end{figure}

Experiments use an AgileX LIMO differential-drive robot with an Astra/Dabai U3 RGB-D camera, 2D LiDAR, and NVIDIA Jetson Orin Nano 8 GB. The robot runs ROS Noetic, GMapping, and \texttt{move\_base}. The OVD node runs YOLO-World-S with one queried target class. Final navigation trials use ``bottle'' and ``cup'' queries to test scheduling across two targets.

\subsection{Policies Compared}

We compare four perception policies:
\begin{itemize}
    \item \textbf{Base-8Hz}: fixed 8 Hz high-power mode (TRACK).
    \item \textbf{Base-5Hz}: fixed 5 Hz mid-power mode (EXPLORE).
    \item \textbf{Base-2Hz}: fixed 2 Hz low-power INT8 mode (SLEEP).
    \item \textbf{Proposed}: tri-state low-/mid-/high-power scheduling.
\end{itemize}
The paper reports LOW/MID/HIGH as conceptual modes and keeps SLEEP/EXPLORE/TRACK as labels where needed. Achieved rates can be below requested values because detector latency and gating limit calls, so results report measured invocation rates.

\subsection{Mission Protocol}


The real-robot evaluation comprises 40 trials spanning two target objects, four perception policies, and five repetitions for each object--policy combination. As shown in Fig.~\ref{fig:exp_setup}, all trials begin from a fixed robot pose, with each target placed at one of five predefined locations. The same placements are reused across policies to support controlled comparison. Bottle and cup trials are conducted in separate batches; between trials, the target is moved to the next location and the navigation stack is reinitialized.

Each trial has a maximum duration of 180~s. Experiments are executed in policy-specific blocks rather than randomized order, so gradual changes in battery state, illumination, or system temperature may introduce temporal drift. Each policy is therefore evaluated across all five placements, and aggregate statistics are reported to reduce sensitivity to individual trial conditions. For every trial, the navigation stack is restarted, system power is monitored externally using \texttt{tegrastats}, and mission outcomes are recorded by the energy-aware navigation framework. The mission-level \texttt{success} field in the corresponding \texttt{mission\_summary} record is used for evaluation.

In addition to the navigation trials, we perform a 60~s stationary microbenchmark for each policy to characterize perception-related power consumption independently of robot motion and navigation workload. During this benchmark, the camera, detector or scheduler, and power logger remain active, while navigation, SLAM, \texttt{move\_base}, and velocity-command execution are disabled. For the proposed policy, scripted inputs exercise the scheduler across low-, mid-, and high-response phases, whereas each fixed-rate baseline remains in its assigned operating mode. This benchmark isolates the contribution of detector execution rate and scheduler state composition from variability introduced by the navigation stack.

\subsection{Metrics and Statistics}

We use the metrics in Section~\ref{sec:methodology_metrics}. Navigation results cover all 40 trials, with aggregate summaries over bottle and cup. TTD is averaged over successful runs only, while Jetson energy and runtime are averaged over all runs because failures consume real time and power. The stationary benchmark is reported separately as a mechanism check. With five placements per object-policy pair, we report descriptive rather than inferential statistics.

\section{Results and Discussion}

\subsection{Navigation Results}

\begin{table}[ht]
\centering
\scriptsize
\setlength{\tabcolsep}{1.4pt}
\caption{Real-robot navigation summary over bottle and cup trials (ten trials per policy). low-, mid-, and high-power modes.}
\label{tab:main_nav_results}
\begin{tabular}{@{}lccccccc@{}}
\toprule
\textbf{Policy} & \textbf{Succ.} & \textbf{$\bar{P}_{J}$} & \textbf{$E_{\mathrm{m}}$} & \textbf{Run} & \textbf{TTD$_{\mathrm{succ}}$} & \textbf{Hz} & \textbf{L/M/H}\\
 & & \textbf{(W)} & \textbf{(J)} & \textbf{(s)} & \textbf{(s)} & & \textbf{(\%)}\\
\midrule
Base-8Hz & 7/10 & 9.35 & 866.9 & 92.8 & 25.9 & 4.74 & 0/0/100\\
Base-5Hz & 7/10 & 8.92 & 1019.8 & 114.2 & 53.6 & 3.39 & 0/100/0\\
Base-2Hz & 7/10 & 8.38 & 1155.0 & 140.4 & 0.9 & 1.48 & 100/0/0\\
Proposed & 8/10 & 8.86 & 1114.0 & 125.9 & 10.9 & 3.12 & 25/60/15\\
\bottomrule
\end{tabular}
\begin{minipage}{0.98\columnwidth}
\vspace{0.5mm}
\scriptsize\emph{Note:} $E_{\mathrm{m}}$ is mission-window Jetson input energy; TTD is conditional on success. L/M/H maps raw log labels to low-/mid-/high-power modes.
\end{minipage}
\end{table}

\begin{figure}[ht]
  \centering
  \includegraphics[width=0.98\columnwidth]{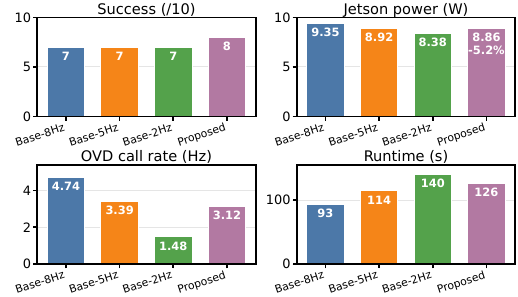}
  \caption{Policy dashboard over the 40 navigation trials. The proposed scheduler has the highest aggregate success count and lower mean Jetson input power than Base-8Hz, while Jetson input energy is dominated by runtime.}
  \label{fig:policy_dashboard}
\end{figure}

\begin{figure}[ht]
  \centering
  \includegraphics[width=0.95\columnwidth]{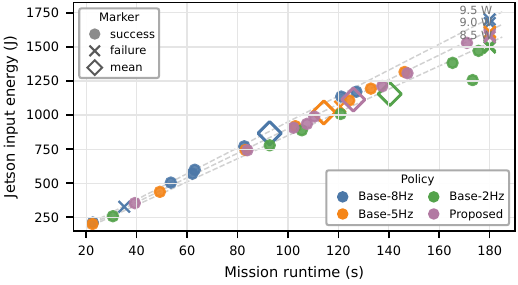}
  \caption{Mission runtime versus Jetson input energy for all 40 trials. Circles are successes, crosses failures, and open diamonds condition means; dashed lines show constant-power contours.}
  \label{fig:nav_tradeoff}
\end{figure}

Table~\ref{tab:main_nav_results} summarizes the 40 real-robot navigation trials. The proposed scheduler has the highest aggregate success count, 8/10, while each fixed-rate baseline succeeds in 7/10 trials. It reduces mean Jetson input power from 9.35 W for Base-8Hz to 8.86 W, a 5.2\% reduction, and lowers achieved OVD rate from 4.74 Hz to 3.12 Hz. Runtime is the mission-window duration over all runs; TTD$_{\mathrm{succ}}$ measures only first accepted detection on successful runs. Base-2Hz has the lowest mean power, 8.38 W, but the longest mean runtime and largest mean Jetson input energy.
Accordingly, the navigation result is best read as power/workload reduction at competitive task reliability, not as a statistically large reliability improvement.

Fig.~\ref{fig:policy_dashboard} visualizes the same summary. The main claim is a power-management trade-off, not guaranteed $E_{\mathrm{m}}$ minimization. Relative to Base-8Hz, the scheduler reduces mean power and detector calls, yet mean $E_{\mathrm{m}}$ is higher because missions last 35.8\% longer on average. Base-2Hz is clearer: it cuts mean power by 10.3\%, but runtime rises by 51.4\%, producing the largest mean $E_{\mathrm{m}}$. Its low TTD$_{\mathrm{succ}}$ reflects successful easy cases, not generally faster search. Fig.~\ref{fig:nav_tradeoff} shows why energy-aware OVD must be evaluated as a closed-loop workload.

\subsection{Time-Domain Scheduler Behavior}

\begin{figure}[ht]
  \centering
  \includegraphics[width=0.95\columnwidth]{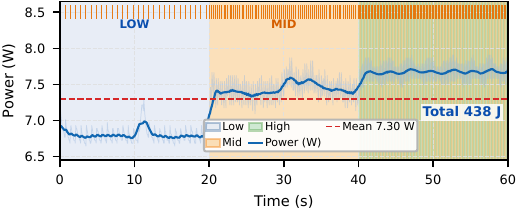}
  \caption{Compact proposed-scheduler trace from the 60 s scripted stationary run. Colored regions show state, orange ticks show detector calls, the blue curve is Jetson input power, and the red dashed line is mean power.}
  \label{fig:trace}
\end{figure}

\begin{figure}[ht]
  \centering
  \includegraphics[width=0.98\columnwidth]{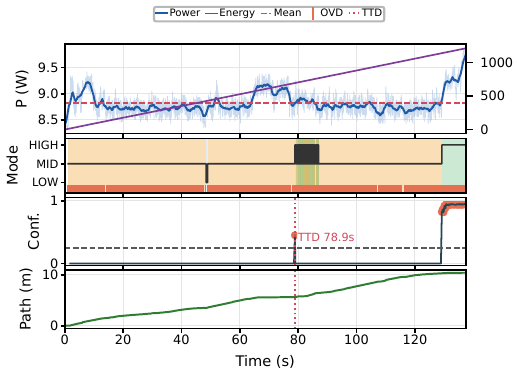}
  \caption{Representative adaptive navigation trace showing Jetson input power/energy, scheduler state and detector calls, detector confidence with the 0.25 threshold, and path length.}
  \label{fig:nav_trace}
\end{figure}

\begin{figure*}[ht]
  \centering
  \includegraphics[width=0.91\textwidth]{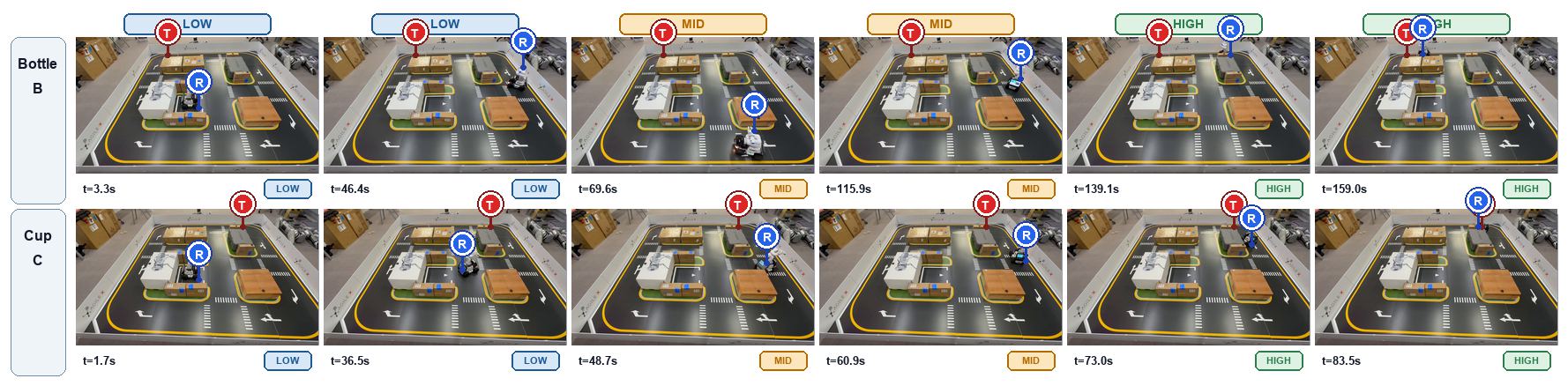}
  \caption{Qualitative video-frame progression for representative bottle and cup searches. Blue R marks the robot and red T marks the target; LOW/MID/HIGH labels are qualitative stage estimates, not synchronized scheduler labels.}
  \label{fig:video_grid}
\end{figure*}

Fig.~\ref{fig:trace} shows the proposed scheduler in the 60 s scripted stationary run. This is not a navigation trial: it isolates detector/scheduler power from path length, obstacle avoidance, and mission duration. The run has 7.30 W mean Jetson input power, consumes 438.1 J, executes 202 detector calls, and spends 50/38/13\% of the horizon in LOW/MID/HIGH modes. The power step between LOW and MID/HIGH makes the mechanism behind Table~\ref{tab:stationary_results} visible in time.

Fig.~\ref{fig:nav_trace} complements the stationary trace with a closed-loop run. It shows the same scheduler while the robot moves, accumulates path length, and receives intermittent target-confidence events. The run is illustrative, exposing timing relationships among power, state, confidence, and path progress.
Fig.~\ref{fig:video_grid} adds visual context from two run videos. Early frames show stable or low-value viewpoints where LOW is appropriate, middle frames show obstacle-aware navigation where MID maintains periodic OVD, and late frames show target-adjacent candidate response where HIGH is appropriate. The figure is qualitative; quantitative occupancy and timing come from synchronized logs in Tables~\ref{tab:main_nav_results} and~\ref{tab:stationary_results} and Fig.~\ref{fig:nav_trace}.

\subsection{Stationary Power Mechanism}

\begin{wraptable}{r}{0.55\columnwidth}
\vspace{-10pt}
\centering
\scriptsize
\setlength{\tabcolsep}{1.7pt}
\caption{Stationary microbenchmark over the power-integration window.}
\label{tab:stationary_results}
\begin{tabular}{@{}lrrrrl@{}}
\toprule
\textbf{Policy} &
\textbf{$\bar{P}_{J}$} &
\textbf{$E$} &
\textbf{Hz} &
\textbf{\#Inf.} &
\textbf{L/M/H} \\
&
\textbf{(W)} &
\textbf{(J)} &
&
&
\textbf{(\%)} \\
\midrule
Base-8Hz & 8.045 & 484.7 & 6.67 & 402 & 0/0/100 \\
Base-5Hz & 7.578 & 456.6 & 4.40 & 265 & 0/100/0 \\
Base-2Hz & 6.847 & 410.8 & 1.90 & 114 & 100/0/0 \\
Proposed & 7.301 & 438.1 & 3.37 & 202 & 50/38/13 \\
\bottomrule
\end{tabular}

\vspace{0.5mm}
\footnotesize
\emph{Note:} OVD Hz is computed as \#Inf./mission-window duration. 
\vspace{-2mm}
\end{wraptable}

The stationary microbenchmark isolates detector and scheduler power from navigation-related variability. OVD frequency is computed from detector invocations within the same time window used for power and energy
integration. The fixed-rate baselines exhibit the expected monotonic reduction in power as the detector rate decreases from Base-8Hz to Base-5Hz and Base-2Hz.

The proposed scheduler operates between the Base-5Hz and Base-2Hz baselines, reflecting its mixed operating profile: approximately 50\% of the benchmark is spent in LOW, 37.6\% in MID, and 12.5\% in HIGH. Relative to Base-8Hz, the proposed policy reduces mean power by 9.2\% and energy consumption over the 60~s benchmark by 9.6\%.

\subsection{Object-Level Observations}

\begin{figure}[ht]
  \centering
  \includegraphics[width=0.94\columnwidth]{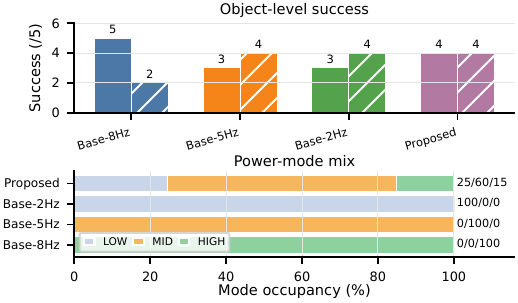}
  \caption{Object-level success and state occupancy, mapped to LOW/MID/HIGH power modes. In the top plot, solid bars denote bottle and hatched bars denote cup.}
  \label{fig:object_state}
\end{figure}

The combined result hides object-level variation. On bottle trials, Base-8Hz succeeds in 5/5 runs, the proposed scheduler in 4/5, and the 5 Hz and 2 Hz baselines in 3/5 each. On cup trials, the proposed scheduler, Base-5Hz, and Base-2Hz each succeed in 4/5, while Base-8Hz succeeds in 2/5. Fig.~\ref{fig:object_state} shows this split and state mix. Thus, the adaptive policy has the highest aggregate success count here, but the sample is too small to claim universal dominance.



\section{Conclusion}

This paper presented an energy-aware OVD scheduler for low-power mobile robots. It treats detector frames as mission resources: low-value frames can be sampled slowly, while candidate-target phases benefit from high-rate response. In 40 real AgileX LIMO navigation trials across bottle and cup searches, the scheduler shows 8/10 success compared with 7/10 for each fixed-rate baseline and reduces mean Jetson input power by 5.2\% relative to Base-8Hz. The same trials show the key caution: lower mean power does not guarantee lower mission-window Jetson energy, because longer runtime can dominate the integral. A stationary microbenchmark confirms the mechanism, with 9.6\% lower 60 s energy than Base-8Hz. Future work will broaden objects and environments, add external labels, randomize trials, measure whole-robot energy, and study motion-predictive scheduling.

\begingroup
\footnotesize
\bibliographystyle{IEEEtran}
\bibliography{cite}
\endgroup

\end{document}